# Hyper-Heuristic Algorithm for
# Finding Efficient Features in Diagnose of Lung Cancer Disease

Mitra Montazeri[1,2], MahdiehSoleymani Baghshah[3], Ahmad Enhesari[4, a]

[1]Medical Informatics Research Center, Institute for Futures Studies in Health, Kerman University of Medical Sciences, Kerman, Iran.
[2]Computer Engineering Department, Shahid Bahonar University, Kerman, Iran,
[3]Computer Engineering Department, Sharif University of Technology, Tehran, Iran.
[4]Department of the radiology, Kerman University of Medical Sciences, Kerman, Iran.

**ABSTRACT**

*Background*: Lung cancer was known as primary cancers and the survival rate of cancer is about 15%. Early detection of lung cancer is the leading factor in survival rate. All symptoms (features) of lung cancer do not appear until the cancer spreads to other areas. It needs an accurate early detection of lung cancer, for increasing the survival rate. For accurate detection, it need characterizes efficient features and delete redundancy features among all features.Feature selection is the problem of selecting informative features among all features.

*Materialsand Methods*: Lung cancer database consist of 32 patient records with 57 features. This database collected by Hong and Youngand indexed in the University of California Irvine repository. Experimental contents include the extracted from the clinical data and X-ray data, etc. The data described 3 types of pathological lung cancers and all features are taking an integer value 0-3. In our study, new method is proposed for identify efficient features of lung cancer. It is based on Hyper-Heuristic.

*Results*:We obtained an accuracy of 80.63% using reduced 11 feature set.The proposed method compare to the accuracy of 5 machine learning feature selections.The accuracy of these 5 methods are 60.94, 57.81, 68.75, 60.94 and 68.75.

*Conclusions*: The proposed method has better performance with the highest level of accuracy. Therefore, the proposed model is recommended for identifying an efficient symptom of Disease. These finding are very important in health research, particularly in allocation of medical resources for patients who predicted as high-risks

## 1  INTRODUCTION

Lung cancer is a deadly cancer that has killed large number of people every year. The survival rate of cancer is about 15%. Statistics show that victims of breast, prostate and colon overall are less than the lung cancer victims. Approximately 60% of patients are men and 33% of women have lung cancer. Lung cancer is characterized by uncontrolled cell growth in tissues of the lung. If not diagnosed, this growth can spread beyond the lung into nearby tissue in a process called metastasis and gradually, into other parts of the body.

Due to most cancers start in lung, they are known as primary cancers which they are carcinomas that derive from epithelial cells. 80–90% of lung cancers [1] in reason of long-term exposure to tobacco smoke,[2] these cases are often attributed to a combination of genetic factors,[3]radon gas,[3]asbestos,[4] and air pollution[4] including secondhand smoke while nonsmokers account is only 10–15% of lung cancer cases.[5]

Symptoms of lung cancer are coughing up blood. It is the most common symptoms in lung cancer. Weight loss and shortness of breath [1] can be another symptoms of this disease. Lung cancer may be seen on chest radiograph and computed tomography (CT scan).

With a biopsybiopsy the diagnosis is confirmed[6]**.**Bronchoscopy or CT-guided biopsy performs biopsybiopsy. Common treatments include surgery, chemotherapy, and radiotherapy. Treatment and prognosis depend on the histological type of cancer, the stage (degree of spread), and the patient's general well-being, measured by performance status.

Survival depends on stage, overall health, and other factors. Overall, 15% of people in the United States diagnosed with lung cancer survive five years after the diagnosis.[7] Worldwide, lung cancer is the most common cause of cancer-related death in men and women, and is responsible for 1.38 million deaths annually, as of 2008.[8]. However the 5-

---

[a]**Corresponding author**: Ahmad Enhesari, Department of the radiology, Kerman University of Medical Sciences, Kerman, Iran. Email:enhesari@kmu.ac.ir.

year survival rate of localized stage is about 50%. Localized stage cancer is the cancer that does not spread to additional sites like lymph node within the body. Early detection of lung cancer is the leading factor in survival rate.

Due to have so many factors to diagnose the lungcancer patient. It causes to make the physician's job difficult.A physician often makes decisions by evaluating the current test results of a patient and by referring to the previousdecisions he made on other patience with the same condition [9]. On the other, all symptoms of lung cancer do not appear until the cancer spreads to other areas, thus leading to cancer detection of only 24% in early stages [10, 11]. We need an accurate early detection of lung cancer, for increasing the survival rate. For accurate detection of this disease, it need characterizes efficient features (symptoms) and delete redundancy or unrelevant feature among all features.

Features are useful in many aspect of applications [31-33] and feature selection is the problem of selecting informative features among all features such that the selected feature subset has lower cardinality and provides higher accuracy.

Until now, researchers have studied various aspects of feature selection. There are two key aspects of feature selection: feature evaluation and search strategies. Feature evaluation is about how to measure the goodness of a feature subset [12, 13].

In search strategies, heuristic searches employ heuristics to conduct the search process. Due to their polynomial complexity, they can be implemented much faster than the previous search strategies.Until now many heuristic algorithms with different strategies have been proposed [14], but they have their own problem.Hyper-heuristic (HH) approach is a newest heuristic algorithm which was introduced in 2000 [23], solve these problems, properly. It has two levels, at the low level, there is a set of local searches which also known as Low Level Heuristics (LLH)s. The characteristic of these LLHs may be problem-oriented and may change from one problem to another. At the high level, there is a black-box choice function. It is a supervisor that manages the choice of low level heuristic. Recent studies on HHs have verified their successes on a wide variety of real-world problems. They not only converge to high-quality solutions but also search more efficiently than their conventional counterparts [15-17].

In this paper, efficient feature of lung cancer is identified by Hyper-heuristic. The goal of our method is to improve classification performance and search to identify important feature subsets. Proposed method is compared by most common machine learning feature selection.

The rest of this paper is organized as follows: Section 2 describes preliminaries of some definition. In Section 3, we explain the proposed method. Finally, experimental results and conclusion are presented in Section 4 and 5, respectively.

## 2 Preliminaries

In this section, at first some basic concepts about Pearson Correlation Coefficient (as we use it later as the evaluation criterion) are given, and then the concept of feature selection.

### 2.1 Pearson Correlation Coefficient

Correlation coefficients are one of the simplest approaches to feature relevance measurements. In contrast to information theoretic and decision tree approaches, they avoid problems with probability density estimation and discretization of continuous features and therefore they are treated first [12].

The linear correlation coefficient of Pearson is very popular in statistics and represents the normalized measure of the strength of linear relationship between variables [18]. For random variable X with values x and random variable Y with values y, while a vector of $d$ data points $x_i, y_i$, $i = 1,\ldots, d$, it is defined as:

$$r_{x,y} = \frac{\sum_i (x_i - \bar{x}_i)(y_i - \bar{y}_i)}{\sqrt{\sum_i (x_i - \bar{x}_i)^2 \sum_j (y_i - \bar{y}_i)^2}}. \tag{1}$$

Where, as usual, $\bar{x}$ is the mean of the $x_i$'s, $\bar{y}$ is the mean of the $y_i$'s. $r_{xy}$ is equal to ± 1 if *X* and *Y* are linearly dependent and zero if they are completely uncorrelated (Random variables may be correlated positively or negatively).

If a group of k features has already been existed, correlation coefficients may be used to estimate correlation between this group and the class variable, including inter-correlations between features. Relevance of a group of features grows with the correlation between features and classes, and decreases with growing inter-correlation. These ideas have been discussed in the theory of psychological measurements [19] and in the literature on decision making and aggregating opinions [20]. In 1964, Ghiselli [19] proposed the following equation:

$$J_k = \frac{k\overline{r}_{cf}}{\sqrt{k + k(k-1)\overline{r}_{ff}}}. \tag{2}$$

Where $\overline{r}_{cf}$ is the average correlation coefficient between these $k$ features and the output variables and the average between different features as $\overline{r}_{ff}$. This formula is obtained from Pearson's correlation coefficient where all variables have been standardized. It has been used in the Correlation-based Feature Selection (CFS) algorithm [21].

### 2.2 Feature Selection Problem

We can formulate the problem of selecting a subset of features with superior classificatory performance as follows: Let $F$ be the original set of features with cardinality $N$ and $m$ be the number of features in the selected subset, $X$, $X \subseteq F$. Let the feature selection criterion function for the set $X$ be represented by $J(X)$ (higher value of $J$ indicates better feature subset). Formally, the problem of feature selection is to find a subset $X \subseteq F$ which has the following two important properties:
- $m < N$
- $J(X) > J(N)$

It has been demonstrated that searching for the minimum feature subset is NP-hard [22].

## 3 Proposed Method

The proposed method chooses LLHs appropriately by using GA strategies. Since each region of the solution space can have its own characteristics, an appropriate LLH should be selected and applied to the current solution. GA's role is a supervisor which manages the choice of LLHs that should be applied at any time. GA chooses a LLH based on the existing functional history of LLHs. This GA is not a direct GA, in fact each individual in GA's population consists of a sequence of integer numbers. Each number is an LLH choice which tells us which LLH must be applied and each individual tells in which order to apply LLHs.

The LLHs perform local searches on current solution to improve it. Local search is a neighborhood searching algorithm. They cooperate in each chromosome by exchanging their local best solutions in order to combine their efforts and to improve the quality of the solutions that each of them would be able to find by itself (working on a standalone basis). Indeed, we use the domain knowledge of the feature selection problem in designing the LLHs to provide more effective search. The local best solutions are the complete set of best solutions found by the LLHs at each chromosome. A set of LLHs consists of two different types of local searches: exploiters LLHs and Explorer LLHs. Exploiters LLHs improve the quality of the candidate solution at hand. The aim of an exploiter is to produce a better candidate solution at each step. They cause to increase exploitation while explorer heuristics are not expected to produce a better candidate solution after they are applied. Explorer heuristics dwell on random perturbation and cause to increase exploration.

### 3.1 Chromosome encoding

In our proposed GA, each chromosome $C_i$ consists of a sequence of integer numbers where each number is an LLH choice telling us which LLH must be applied. Each integer is lied in the interval 1 to number of LLHs (*NLLH*). Fig. 1 shows the encoding representation of a chromosome for *NLLH*=12. These integers are generated randomly. As you can see in Fig. 1, for example integer 10 causes to call the 10th LLH.

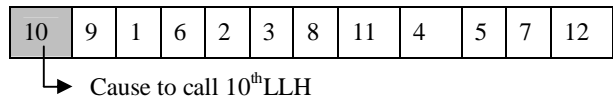

Fig.1: Example of encoding representation of a chromosome as an integer string in the proposed method.

### 3.2 Fitness Function

To evaluate the goodness of each feature subset generated by a chromosome, i.e. sequence of LLHs, we use the accuracy of classification. It can be defined as:

$$Fit = J(fs) \tag{3}$$

Where *fs* denotes the corresponding selected feature subset encoded in the solution and *J* computes the goodness of the feature subset. In our method, *J* shows the accuracy of classification. Note that if two solutions have the same fitness, each one which has smaller number of features is selected.

### 3.3 Evolutionary Operations
Genetic operations are very important to the success of GA applications [24]. In the following, we will present the genetic operations used in the proposed method.

– **Selection**: In the proposed method, the selection operation is based on the fitness of chromosomes. It should ensure that fitter chromosomes have better chance to survive. In our method, we use a rank based elitism roulette wheel selection. Assuming the population has $Np$ chromosomes, for each chromosome $C_i$, $1 \leq i \leq Np$, the selection probability, $p(C_i)$, is calculated as:

$$p(C_i) = \frac{Fit(C_i)}{\sum_{j=1}^{Np} Fit(C_j)} \quad (4)$$

A chromosome $C_i$ is selected if it satisfies the following inequality:

$$\sum_{j=0}^{i-1} p(C_i) < r \leq \sum_{j=0}^{i} p(C_i) \quad (5)$$

Where r is a random number with uniform distribution in [0, 1].

– **Crossover**: We use one-point crossover which if two parent chromosomes $C_1$ and $C_2$ are selected, crossover operation is performed with a crossover probability $P_c$ to generate two new chromosomes $Off_1$ and $Off_2$ through the way which exchange information in a randomly cut point.

– **Mutation**: Mutation operator selects some positions with a mutation probability $Pm$ in one individual randomly and mutates genes at these positions to other values ranging from 1 to $NLLH$[25].

### 3.4 Low Level Heuristics
In the proposed method, 12 LLHs (*NLLH* is equal to *12*) are used to change the current solution. All of our LLHs are divided into two groups: exploiter heuristics and explorer heuristics. We use the domain knowledge of the feature selection problem in designing the local searches to provide a more effective search through the HH approach. 8 of them which are exploiters, search among existent features, nonexistent and all features. Their searches are similar to hill climber[26, 27]. The last four LLHs try to high diversity of search. In these LLHs different emphasis of probability, i.e., random probability, equal probability and probability of mutation in GA is used. These varieties in probability due to each region of the solution space have its own characteristics, besides we use variety of exploiter LLHs we use variety of explorer LLHs.

#### 3.4.1 Solution Encoding and Evaluation
In the proposed method, the encoded solution used by LLHs is a binary string of length equal to the total number of features. Each of its bits encodes a single feature so that the length of the chromosome is *N*. As shown in Fig 2, "1" bit implies the corresponding feature is selected and "0" implies it is excluded. To provide an efficient and also effective method, we use the criteria in Eq. (2) as the evaluation criteria.

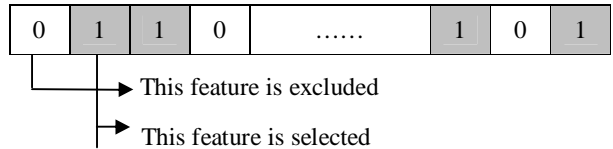

Fig 2: Solution encoding as a binary bit string.

## 4 Experimental Result and Discussion
In this section, several experiments are conducted to evaluate the performance of our method. In the following subsections, brief description of lung cancer dataset and experimental setup will be firstly introduced, and then our results will be presented and compared to results of some literature works.

### 4.1 Database Description and preprocessing

Lung cancer database consist of 32 patient records with 57 features. This database collected by Hong and Young [29] and indexed in the University of California Irvine (UCI) repository [30]. Experimentalcontentsincludethe extractedfrom the clinical data and X-ray data, etc. It should be noted that the numberof samples givenisfewer than the numberof feature. The data described 3 types of pathological lung cancers andall features are nominal, taking an integer value 0-3. Since this dataset has missing values or continues values in uncontrolled ranges, they need a preprocessing step before they are used. For missing values, we replaced them with the most frequently used values for nominal and means of values for numeric features. To control the range of continues features, we normalize them in the range [0, 1].

### 4.2 Performance Evaluation Setup

In our method, the following metrics are used:

- **Result validation**: We use accuracy of 1 nearest neighborhood classifier (1NN) for result validation. It is stable and not sensitive to the initial setting. Other classifiers like neural networks due to producing different output by changing the initial weights may not be proper for performance comparison.
- **k-Fold Cross Validation**: It is used to estimate how accurately a predictive model will perform in practice and represents the probability that an instance is classified correctly. In $k$-fold cross-validation, the original sample is randomly partitioned into $k$ subsamples. A single subsample is retained as the test data and the remaining $k-1$ subsamples are used as training data. The cross-validation process is then repeated $k$ times, with each of the $k$ subsamples used exactly once as the validation data. The $k$ results from the folds then can be averaged to produce a single estimation. In our study, due to being randomness, run 10 times and at each time a 10-fold cross validation which is commonly used [28] is used, and the final results were their average values (10-10 fold CV).

### 4.3 Performance of the proposed method

In this section, we present an experimental study of the proposed method on Lung cancer database. We set the population size $(Np)$, to 30 and the number of generations to 200. Crossover rate $(Pc)$ and mutation rate $(Pm)$ is set to 0.7 and 0.1, respectively. Table 1 shows the best and average accuracy of 10 runs of our method on Lung cancer database. Because the proposed method is a random search algorithm, different results may be obtained at different runs. Therefore, we run this algorithm 10 times and report their average. In addition to the accuracy, the number of selected features is also reported. As shown in this table, this number decreases considerably and for low number of features we can see higher accuracy. Our method can achieve these excellent effects in low generations (200). In fact, heuristic algorithms should be run for high generation, e.g. at least 10000, to find global optimum but our method find their results in 200 generations.

Table 1 performance and number of selected features of the proposed method (1NN, 10-10 folds CV, Unit: %)

| Database | Unselected features | Best results | Best number of selected features | Average results | Mean number of selected features |
|---|---|---|---|---|---|
| Lung Cancer | 56 | 80.63 | 11 | 0.7506 | 18.8 |

### 4.4 Comparison Of literature Works

We compare the proposed method to 5most common machine learning feature selections (for more information refer to Ref. [34-38]) which are listed in Table 2.We compare the average and the best value among 10 runs of the proposed method. According to the results shown in Table 2, in all, our method shows the best results (on both average and best results on 10 runs). Moreover, the achieved results by the proposed method are considerably better than the other methods. Although two methods have lower features but the accuracies of them are considerably lower than the proposed method. In prediction and classification main issue is accuracy and number of features is the second important issue.

About proposed method, reported value is the best value among 10 runs, the value in parenthesis is average value of these 10 runs and average number of the selected features.

Table 2: The comparative results of the proposed method with machine learning feature selector (Unit: %).

| Method | Classification accuracy (%) | Number of selected features |
|---|---|---|
| Gain-Ratio Algorithm | 60.94 | 21 |
| Principal-Components Algorithm | 57.81 | 25 |
| ReliefF Algorithm | 68.75 | 5 |
| Symmetrical-Uncert Algorithm | 60.94 | 12 |
| Chi-Squared Algorithm | 68.75 | 7 |

| The proposed method | **80.63(75.06)** | 11(18.8) |

Due to Chi-Squared algorithm, Gain-Ratio algorithm, ReliefF algorithm and Symmetrical-Uncert algorithm rank features. For gaining feature subset optimum, it is needed to compute classification accuracy of *N* subsets. These *N* subsets are generated by adding ranked feature at *N* step. Classification accuracy of these *N* subset and optimum subset are shown in Fig.3.

**5. Conclusion and Discussion**
Lung cancer is a deadly cancer which has killed large number of people every year. The survival rate of cancer is about 15%.Statistics show that victims of breast, prostate and colon overall are less than the lung cancer victims. If not diagnosed, this growth can spread beyond the lung into nearby tissue in a process called metastasis and gradually, into other parts of the body.Early detection of lung cancer is the leading factor in survival rate.
Due to have so many factors to diagnose the lungcancer patient. It causes to make the physician's job difficult.A physician often makes decisions by evaluating the current test results of a patient and by referring to the previousdecisions he made on other patience with the same condition.It needs an accurate early detection of lung cancer, for increasing the survival rate. For accurate detection, it need characterizes efficient features and delete redundancy features among all features. Feature selection is the problem of selecting informative features among all features.
In this paper, a new method was proposed which was based on a new heuristic, Hyper-Heuristic, to find an efficient factors in diagnosing lung cancer. This method could search the solution space effectively by incorporating exploration and exploitation appropriately on low generations (200). The proposed method could trade of between exploitation and exploration for the feature selection problem. In this method, exploitation was done by exploiter heuristics and exploration was done by explorer heuristics. In the other words, exploiter heuristics improve the quality of the candidate solution at hand to produce a better candidate solution at each step while explorer heuristics dwell on random perturbation and cause to increase exploration (not necessarily produce a better candidate solution). The proposed method chooses the local searches appropriately by a GA and therefore it could trade of between exploitation and exploration. Moreover, since intensive exploiter can trap the searching algorithm into a local optimum we study on the number of iterations (in the exploiter heuristics) to control this issue.
The proposed method compared by 5 most common machine learning algorithms. Their results showed that proposed method had better performance. It obtained an accuracy of 80.63% using reduced 11 feature set.

Fig 3: the process of finding feature subset optimum in 4 machine learning feature selections.

Chi-Squared Algorithm              Gain-Ratio Algorithm

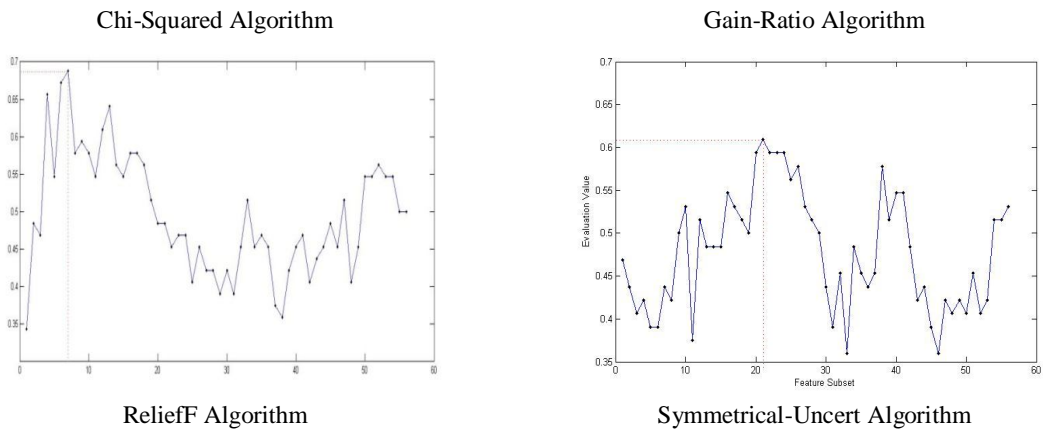

ReliefF Algorithm              Symmetrical-Uncert Algorithm

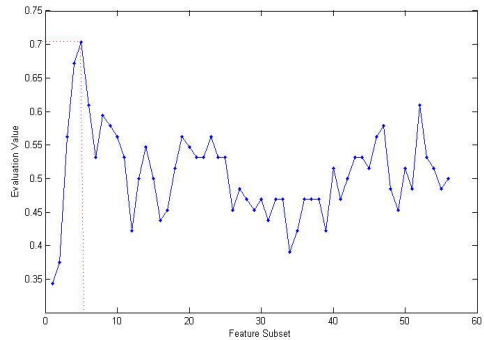 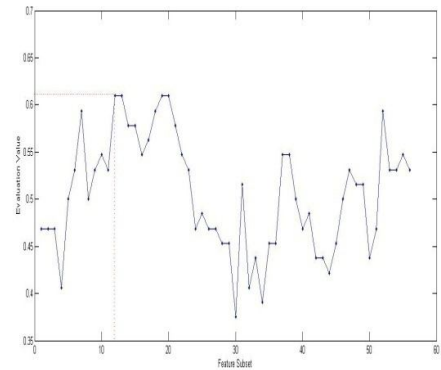